\documentclass[11pt]{article}

\usepackage[margin=1in]{geometry}
\usepackage{booktabs}
\usepackage{tabularx}
\usepackage{array}
\usepackage{amsmath}
\usepackage{amssymb}
\usepackage{graphicx}
\usepackage{xcolor}
\usepackage{hyperref}
\usepackage[numbers,sort&compress]{natbib}

\hypersetup{
  colorlinks=true,
  linkcolor=blue,
  citecolor=blue,
  urlcolor=blue
}

\newcommand{\model}{L20-Edu-135M}
\newcommand{\mean}{\mathrm{Mean}_6}
\newcolumntype{Y}{>{\raggedright\arraybackslash}X}
\newcolumntype{R}{>{\raggedleft\arraybackslash}p{0.09\linewidth}}

\title{\textbf{L20-Edu-135M: An Auditable Single-GPU Study of Data-Efficient Small Language Modeling}}

\author{
  Yin Li\\
  University of Birmingham\\
  \texttt{AliceYin/l20-edu-135m}
}

\date{June 2026}

\begin{document}
\maketitle

\begin{abstract}
Small language models are cheap to serve and feasible on local hardware, but strong public 135M-class systems are commonly trained with hundreds of billions to trillions of tokens on large clusters. We study a sharply resource-constrained regime: a complete 134.5M-parameter language-model pipeline executed on one NVIDIA L20 GPU. The released checkpoint, \model, receives approximately 13B pretraining tokens: 10B FineWeb-Edu tokens followed by a 3B-token educational, mathematics, code, and reasoning mixture. We document the architecture, data gates, cross-source MinHash/LSH near-deduplication, segment deduplication, benchmark-overlap removal, throughput optimization, supervised fine-tuning (SFT) with weight interpolation, and reinforcement learning from verifiable rewards (RLVR) on GSM8K.

In a self-run zero-shot six-task harness, \model\ obtains a mean score of 0.4150. It trails SmolLM-135M (0.4767) and SmolLM2-135M (0.4917), but its mean is 87.1\% of SmolLM-135M's while its nominal token count is 2.17\% as large. This ratio is descriptive, not evidence of statistical equivalence or a controlled scaling law. The model exceeds several older 100M--160M public baselines under the same harness. Direct GRPO-style RLVR decreases GSM8K exact-match accuracy from 1.82\% to 1.59\% (192-token completions) and 1.21\% (320-token completions). These single-run results identify a concrete failure mode rather than establishing a general lower bound on RLVR. The contribution is an auditable resource-constrained case study, not a state-of-the-art claim.
\end{abstract}

\section{Introduction}

The recent small-language-model literature has shown that compact models can become surprisingly capable when trained on carefully curated data and heavily overtrained relative to classical compute-optimal prescriptions. SmolLM and SmolLM2 are prominent examples: their public cards report 600B and 2T pretraining tokens, respectively, for 135M-parameter checkpoints, trained on 64 H100 GPUs \citep{smollm_blog,smollm135_card,smollm2_card,allal2025smollm2}. These models are strong and practical, but their training budgets remain inaccessible to many independent researchers.

This work asks a narrower and more operational question:

\begin{quote}
Can a single commodity data-center GPU support a complete, auditable, competitive 135M-class pretraining and post-training research loop, and what breaks first?
\end{quote}

We answer with a case study around \model, a 134,515,008-parameter Llama-style decoder trained from scratch and then improved through curated continued pretraining and conservative instruction tuning. We organize the study around three research questions:
\begin{description}
  \item[RQ1: Feasibility.] What quality and throughput are attainable for a 135M model when the entire training loop is restricted to one 48GB L20?
  \item[RQ2: Regression control.] Can weight interpolation preserve base-model benchmark performance after instruction tuning?
  \item[RQ3: Small-scale RLVR.] Does exact-answer RLVR improve mathematical reasoning when the 135M starting model rarely solves GSM8K?
\end{description}
The project is framed as a systems and data-efficiency study rather than a leaderboard claim. Its evidence consists of model and data specifications, logged release gates, same-harness baseline evaluations, throughput ablations, regression-gated SFT, and a recorded post-training failure mode.

The central empirical observation is that a 13B-token single-L20 run can become competitive with older 100M-160M public baselines and retain a large fraction of the same-size SmolLM score, but it remains clearly behind modern 135M models trained with 46x to 154x more tokens. This gap is important: it separates what is achievable through careful engineering from what still requires scale.

\paragraph{Contributions.}
We make five concrete contributions.

\begin{enumerate}
  \item We document a single-GPU 135M training recipe: a deep-thin architecture, tied embeddings, grouped-query attention, 10B-token FineWeb-Edu pretraining, and 3B-token curated continued pretraining.
  \item We implement a strict Stage 4 data gate combining cross-source MinHash/LSH near-deduplication, segment-level deduplication, quality filters, and benchmark contamination checks using 13-gram candidates plus LCS overlap rejection.
  \item We evaluate SFT weight interpolation as a regression-control mechanism. Among six tested checkpoints, the selected checkpoint uses 87.5\% SFT weights and 12.5\% Stage 4 base weights; the observed improvement is small and is not claimed to be statistically significant.
  \item We report same-protocol six-task results against public 100M-1B baselines, including SmolLM-135M, SmolLM2-135M, Qwen2.5-0.5B, and OLMo-1B.
  \item We provide a negative RLVR result on GSM8K: direct GRPO and longer completions reduce exact-answer accuracy at 135M, motivating cold-start, distillation, and scaling studies rather than blind RL transfer from large reasoning models.
\end{enumerate}

\section{Related Work}

\paragraph{Scaling laws and overtraining.}
Chinchilla-style scaling laws emphasize joint scaling of model parameters and tokens under a fixed compute budget \citep{hoffmann2022chinchilla}. Later work argues that inference-aware deployment can favor smaller models trained longer than the original training-compute optimum \citep{sardana2024beyondchinchilla}. This matters for 135M models: once inference cost and local deployment are primary constraints, spending more training tokens on a small model can be rational even when it is not classically compute-optimal.

\paragraph{Data-centric language modeling.}
Recent open-data efforts show that data curation can dominate architecture differences at fixed scale. FineWeb documents large-scale Common Crawl filtering and introduces FineWeb-Edu, an educational subset that improves knowledge and reasoning benchmarks \citep{penedo2024fineweb}. DataComp-LM provides controlled dataset experiments and finds model-based filtering crucial for strong models \citep{li2024datacomplm}. Dolma emphasizes transparent corpus documentation \citep{soldaini2024dolma}, while RefinedWeb shows that filtered and deduplicated web data can be highly competitive \citep{penedo2023refinedweb}. Deduplication itself is not merely storage hygiene: it reduces memorization and improves training efficiency \citep{lee2021deduplicating}.

\paragraph{Small model architecture.}
Older 100M-class public baselines such as GPT-2 Small, OPT-125M, GPT-Neo-125M, and Pythia-160M generally use shallower and wider designs \citep{radford2019gpt2,zhang2022opt,biderman2023pythia}. Modern compact models increasingly favor deep-thin designs, tied embeddings, grouped-query attention, and carefully tuned tokenizers. SmolLM explicitly uses a deep-thin 135M architecture with GQA and 600B training tokens \citep{smollm_blog}. SmolLM2 extends this data-centric recipe and reports 2T tokens for the 135M base checkpoint \citep{smollm2_card,allal2025smollm2}.

\paragraph{RLVR and reasoning.}
Reinforcement learning from verifiable rewards became a central post-training direction after DeepSeek-R1, which used rule-based rewards and GRPO-style optimization to elicit reasoning behavior in large models \citep{deepseek2025r1}. However, most reported successes operate at much larger parameter scales or rely on strong cold-start data. Our 135M GSM8K results support the view that verifiable reward alone is not sufficient when the base model rarely solves the task.

\section{System Overview}

\subsection{Model}

\model\ is a 134.5M-parameter decoder-only Transformer. The architecture follows a deep-thin small-model design rather than a GPT-2-style shallow-wide design. Table~\ref{tab:architecture} summarizes the released base configuration.

\begin{table}[h]
\centering
\small
\begin{tabular}{ll}
\toprule
Field & Value \\
\midrule
Parameters & 134,515,008 \\
Architecture & Llama-style causal decoder \\
Layers & 30 \\
Hidden size & 576 \\
FFN size & 1536 \\
Attention & 9 query heads, 3 key/value heads \\
Tokenizer & HuggingFaceTB/SmolLM2-135M tokenizer \\
Initial context length & 2048 tokens \\
Stage 4 context length & 8192 tokens \\
Embeddings & Tied input/output embeddings \\
Precision & bfloat16 \\
\bottomrule
\end{tabular}
\caption{Architecture of \model. The deep-thin shape is close to modern 135M compact models and differs from older 12-layer, 768-hidden 100M-class baselines.}
\label{tab:architecture}
\end{table}

\subsection{Training Stages}

The training pipeline has three major phases. Table~\ref{tab:training} records the principal optimization settings needed to interpret the run. All reported training stages are single runs; we do not average across random seeds.

\begin{table}[h]
\centering
\small
\begin{tabularx}{\linewidth}{Yll}
\toprule
Setting & Initial pretraining & Continued pretraining \\
\midrule
Train tokens & 10,001,252,352 & 3,000,000,965 \\
Context length & 2048 & 8192 \\
Precision & bfloat16 & bfloat16 \\
Optimizer & AdamW & AdamW \\
Peak learning rate & $4\times10^{-4}$ & released config \\
Schedule & warmup + cosine & released config \\
Attention backend & PyTorch SDPA & PyTorch SDPA \\
Execution & compile + checkpointing & compile + checkpointing \\
Hardware & 1x NVIDIA L20 48GB & 1x NVIDIA L20 48GB \\
\bottomrule
\end{tabularx}
\caption{Core training settings. Values not recoverable from the immutable run summary are not reconstructed from memory; the released machine-readable configuration is authoritative.}
\label{tab:training}
\end{table}

\subsubsection*{Stage 1: from-scratch educational pretraining}
The initial model is trained on 10,001,252,352 packed tokens from FineWeb-Edu, using streaming data with score thresholding and document length filters. The run uses AdamW, a peak learning rate of \(4\times10^{-4}\), warmup plus cosine decay, bfloat16, PyTorch SDPA, torch.compile, gradient checkpointing, and 528,384 tokens per optimizer step. It completes at step 18,928 with validation loss 2.8731 and perplexity 17.69.

\subsubsection*{Stage 4: curated continued pretraining}
The second major phase adds 3,000,000,965 train tokens and 4,194,398 validation tokens from a curated mixture. The goal is to improve educational, code, mathematics, and reasoning coverage while limiting data repetition and benchmark contamination. Table~\ref{tab:stage4mix} lists the token allocation.

\begin{table}[h]
\centering
\small
\begin{tabularx}{\linewidth}{Y r r}
\toprule
Source & Tokens & Share \\
\midrule
FineWeb-Edu score 3+ & 1,200,000,008 & 40.0\% \\
DCLM educational shards & 900,000,863 & 30.0\% \\
High-quality educational replay & 299,999,994 & 10.0\% \\
Dolmino PES2O & 210,000,316 & 7.0\% \\
Dolmino Wikipedia & 150,000,406 & 5.0\% \\
Dolmino StackExchange & 90,000,106 & 3.0\% \\
FineMath score 4+ & 90,001,567 & 3.0\% \\
MixtureVitae formatted tutorials & 21,001,247 & 0.7\% \\
MixtureVitae reasoning & 20,996,085 & 0.7\% \\
MixtureVitae math word problems & 18,000,373 & 0.6\% \\
\midrule
Total & 3,000,000,965 & 100.0\% \\
\bottomrule
\end{tabularx}
\caption{Stage 4 continued-pretraining mixture. MixtureVitae is deliberately capped and restricted to structured tutorial, reasoning, and math-word-problem subsets because the project treats incomplete cross-source deduplication as a risk.}
\label{tab:stage4mix}
\end{table}

\subsubsection*{Post-training: SFT and interpolation}
The SFT phase uses filtered SmolTalk-style instruction data. A full SFT checkpoint improves interface behavior but slightly reduces the six-task base-LM mean. We therefore evaluate linear weight interpolations between the Stage 4 base and SFT checkpoint:
\[
  \theta_{\alpha} = (1-\alpha)\theta_{\mathrm{base}} + \alpha\theta_{\mathrm{SFT}}.
\]
The selected release is \(\alpha=0.875\), which achieves the highest six-task mean among tested candidates.

\section{Data Quality and Contamination Control}

The Stage 4 data gate is designed around a practical assumption: for small models with limited token budgets, low-quality repetitions and benchmark overlap can consume a nontrivial fraction of useful learning signal. The gate therefore removes low-quality documents before and after segment cleanup, tracks duplicate segments, and rejects benchmark-overlapping content.

\subsection{Deduplication}

The cross-source guard indexes 3,312,229 documents and 34,852,069 segments. Near-duplicate detection uses 64-permutation MinHash with LSH candidate search across sources. The release gate records 52,995,632 LSH bands. Segment-level cleanup removes duplicate sentence or paragraph units before packing. This is intentionally stricter than per-source deduplication because many public mixtures contain overlapping web, educational, and generated content.

\subsection{Quality Filtering}

The Stage 4 preparation process saw 9,653,444 candidate records and kept 2,961,947. Major rejection categories include quality score rejection, very short examples, repeated n-grams, repeated lines, low alphabetic ratio, high digit ratio, boilerplate, and low unique-word counts. Segment cleanup removed 205,560 duplicate segments and 3,951 too-short segments.

These counts describe the implemented gate, not the causal value of each filter. Because no matched ``filter removed'' control was trained, we do not attribute a benchmark gain to any individual cleaning operation.

\subsection{Benchmark Decontamination}

We build a benchmark contamination index from ARC-Challenge, ARC-Easy, HellaSwag, PIQA, LAMBADA OpenAI, and WinoGrande. Candidate training examples are rejected when 13-gram matching finds a candidate overlap and token-level LCS ratio is at least 0.60. The final Stage 4 gate removed 23 benchmark-overlapping examples: 11 linked to ARC-Challenge, 6 to HellaSwag, and 6 to PIQA. The gate passed with no recorded failures.

This audit is stronger than an unverified web mixture, but it is not a proof of zero contamination. It is an engineering control with explicit thresholds and logged rejection counts.

\subsection{Data Governance}

All constituent corpora are referenced by public dataset identifiers and processed locally rather than redistributed as a new raw-text corpus. A release must preserve each upstream source's license, terms, attribution, and permitted-use constraints. The model card and data manifest are the authoritative locations for exact dataset revisions and license metadata. Web-derived corpora can contain personal, copyrighted, toxic, or factually incorrect material even after filtering; the present pipeline does not establish complete removal of such content. Consequently, the checkpoint is a research artifact and should not be treated as a verified factual or safety-critical system.

\section{Training Efficiency}

The project treats tokens per second as a first-class metric because single-GPU feasibility depends on throughput. Table~\ref{tab:speed} summarizes the best observed short-run configurations.

\begin{table}[h]
\centering
\small
\begin{tabularx}{\linewidth}{Y r r r r}
\toprule
Configuration & Context & Micro-batch & Tokens/s & Peak allocated GB \\
\midrule
SDPA + Liger + compile & 2048 & 16 & 49,119 & 25.05 \\
SDPA + Liger + compile & 2048 & 12 & 48,823 & 18.98 \\
SDPA + Liger & 2048 & 16 & 46,965 & 21.80 \\
SDPA + Liger + compile & 4096 & 8 & 34,171 & 28.80 \\
SDPA + Liger + compile & 4096 & 10 & 34,058 & 35.74 \\
SDPA + Liger & 4096 & 12 & 33,581 & 37.86 \\
\bottomrule
\end{tabularx}
\caption{Measured L20 speed ablations. The fastest measured 2048-token setup reaches 49.1k tokens/s; 4096-token training is slower but still practical.}
\label{tab:speed}
\end{table}

The initial 10B run logs a mean throughput of 38,541 tokens/s and 38,587 tokens/s after step 1000. At that rate, the 10B-token run takes about 72 GPU-hours. Later optimized short runs reach approximately 49k tokens/s and about 50.7\% model FLOP utilization under the project's MFU estimator. These are short-run engineering measurements rather than end-to-end training averages. Energy is estimated from observed GPU power around 341--354 W; the logged window corresponds to roughly 2.0 Wh per million tokens for GPU-only energy, excluding host, cooling, storage, networking, setup, failed runs, and idle time.

\section{Evaluation Protocol}

We evaluate with EleutherAI lm-evaluation-harness 0.4.12 using the Hugging Face backend, bfloat16, CUDA, full task datasets, zero-shot settings, logged samples, and a consistent comparison parser. Public baselines are self-run under the same harness rather than copied from leaderboards. The six tasks are ARC-Challenge \citep{clark2018arc}, ARC-Easy \citep{clark2018arc}, HellaSwag \citep{zellers2019hellaswag}, LAMBADA OpenAI \citep{paperno2016lambada}, PIQA \citep{bisk2019piqa}, and WinoGrande \citep{sakaguchi2019winogrande}. Their evaluation sets contain 1,172, 2,376, 10,042, 5,153, 1,838, and 1,267 examples, respectively.

The primary scalar metric is the arithmetic mean over the selected metric for each task:
\[
  \mean = \frac{1}{6} \sum_{t=1}^{6} s_t .
\]
For ARC-Challenge, ARC-Easy, HellaSwag, and PIQA, the selected metric is normalized accuracy. For LAMBADA OpenAI and WinoGrande, it is accuracy.

This is a public-checkpoint comparison, not a tokenizer-controlled architecture experiment. Models use their own tokenizers, context lengths, and pretraining histories. The comparison is useful for external reference but cannot isolate architecture from data or tokenizer effects.

\paragraph{Selection and uncertainty.}
The six-task mean was used to select among the six SFT interpolation candidates, so the selected mean is subject to selection bias. Baseline model choices were fixed by parameter scale and public availability. Full evaluation sets remove test-subsampling variance but not training-seed, checkpoint-selection, implementation, or dataset uncertainty. Since each model was evaluated once and no independent training replicas exist, we do not attach seed-based error bars or claim significance for small score differences. In particular, the 0.0009 increase from the Stage 4 base to the selected interpolation is descriptive only.

\section{Results}

\subsection{Six-Task Baselines}

Table~\ref{tab:benchmarks} gives the main benchmark comparison. \model\ reaches a six-task mean of 0.4150. It is ahead of GPT-2 Small, OPT-125M, Pythia-160M, and Cerebras-GPT-111M in mean score, but remains behind SmolLM-135M and SmolLM2-135M.

\begin{table}[h]
\centering
\scriptsize
\begin{tabular}{lrrrrrrr}
\toprule
Model & ARC-C & ARC-E & HellaSwag & LAMBADA & PIQA & WinoGrande & \(\mean\) \\
\midrule
SmolLM2-135M & 0.2969 & 0.5854 & 0.4301 & 0.4289 & 0.6839 & 0.5249 & 0.4917 \\
SmolLM-135M & 0.2875 & 0.5610 & 0.4265 & 0.3757 & 0.6823 & 0.5272 & 0.4767 \\
\model\ final & 0.2867 & 0.4958 & 0.3240 & 0.2602 & 0.6148 & 0.5083 & 0.4150 \\
OPT-125M & 0.2210 & 0.3990 & 0.3160 & 0.3856 & 0.6202 & 0.5178 & 0.4099 \\
GPT-2 Small & 0.2261 & 0.3973 & 0.3138 & 0.3076 & 0.6208 & 0.5067 & 0.3954 \\
Pythia-160M & 0.2312 & 0.3641 & 0.3030 & 0.1225 & 0.5979 & 0.5075 & 0.3544 \\
Cerebras-GPT-111M & 0.2099 & 0.3506 & 0.2720 & 0.1912 & 0.5811 & 0.4901 & 0.3491 \\
\bottomrule
\end{tabular}
\caption{Zero-shot six-task benchmark comparison under the same self-run harness. \model\ is competitive with older 100M-160M public baselines but remains below modern data-centric 135M models.}
\label{tab:benchmarks}
\end{table}

\subsection{Token-Budget Context}

Table~\ref{tab:budget} compares reported token budgets. \model\ reaches 87.1\% of SmolLM-135M's six-task mean with 13B tokens, or about 2.17\% of the SmolLM-135M reported pretraining budget. It reaches 84.4\% of SmolLM2-135M's six-task mean with about 0.65\% of the reported SmolLM2-135M token budget.

\begin{table}[h]
\centering
\small
\begin{tabularx}{\linewidth}{Y r r r r}
\toprule
Model & Params & Reported tokens & Hardware in public card & \(\mean\) \\
\midrule
\model\ final & 134.5M & 13B & 1x NVIDIA L20 & 0.4150 \\
SmolLM-135M & 135M & 600B & 64x H100 & 0.4767 \\
SmolLM2-135M & 135M & 2T & 64x H100 & 0.4917 \\
Qwen2.5-0.5B & 0.49B & not listed in card & not listed in card & 0.5363 \\
OLMo-1B & 1B & 3T & not listed in HF card & 0.5681 \\
\bottomrule
\end{tabularx}
\caption{Training-budget context. Token counts and hardware for SmolLM and SmolLM2 come from public model cards. Qwen2.5-0.5B and OLMo-1B are larger reference checkpoints, not same-size baselines.}
\label{tab:budget}
\end{table}

This budget table is the most favorable lens for \model, but it should not be overinterpreted. The ratios divide heterogeneous aggregate accuracies and are not a measure of compute efficiency: score is nonlinear in tokens, and public models differ in data, tokenizer, architecture, optimization, and hardware. The supported claim is that the project establishes a competitive accessible-budget operating point, not that it matches modern 135M state of the art or proves a superior scaling curve.

\subsection{SFT Interpolation}

Table~\ref{tab:sft} shows the anti-forgetting interpolation ablation. The Stage 4 base mean is 0.4141. Full SFT lowers the mean to 0.4137. Interpolation recovers and slightly improves the benchmark mean, with \(\alpha=0.875\) selected at 0.4150.

\begin{table}[h]
\centering
\small
\begin{tabular}{lr}
\toprule
Variant & Six-task mean \\
\midrule
Stage 4 SFT, \(\alpha=0.875\) & 0.41497 \\
Stage 4 SFT, \(\alpha=0.500\) & 0.41465 \\
Stage 4 base & 0.41407 \\
Stage 4 SFT, \(\alpha=0.250\) & 0.41384 \\
Full SFT & 0.41374 \\
Stage 4 SFT, \(\alpha=0.750\) & 0.41332 \\
\bottomrule
\end{tabular}
\caption{SFT interpolation ablation. The final release uses \(\alpha=0.875\) because it passes the regression gate and gives the best six-task mean among tested candidates.}
\label{tab:sft}
\end{table}

The observed difference is too small to establish a general benefit. Operationally, however, interpolation is a cheap checkpoint-level regression control: it generated candidates without additional training and prevented release of a checkpoint with a lower measured base-LM mean. Chat or instruction-following quality was not used in this selection and must be assessed separately.

\subsection{RLVR on GSM8K}

We tested a direct RLVR pipeline using GSM8K exact-answer rewards and TRL GRPO-style optimization. The base Stage 4 SFT/interpolated model scores 24/1319 on GSM8K exact-answer evaluation, or 1.82\%. A first GRPO run with 192-token completions falls to 21/1319. A longer 320-token completion run falls further to 16/1319. A GSM8K chain-of-thought SFT warmup alone reaches 21/1319.

\begin{table}[h]
\centering
\small
\begin{tabular}{lrrr}
\toprule
Checkpoint & Correct & Total & Accuracy \\
\midrule
Stage 4 SFT/interpolated base & 24 & 1319 & 0.0182 \\
GRPO, 192-token completion & 21 & 1319 & 0.0159 \\
GRPO, 320-token completion & 16 & 1319 & 0.0121 \\
GSM8K CoT SFT warmup & 21 & 1319 & 0.0159 \\
\bottomrule
\end{tabular}
\caption{GSM8K exact-answer results. Direct RLVR does not improve the 135M model.}
\label{tab:rlvr}
\end{table}

Diagnostics suggest that the RLVR run reduces some repetition indicators but hurts answer extraction and correctness. For the 320-token run, prediction rate drops to 75.6\%, and exact accuracy decreases by 0.61 percentage points relative to the already weak baseline. The results are consistent with a sparse-reward cold-start problem: when the model rarely solves GSM8K, group-relative rewards may provide too little positive signal. This is a hypothesis, not a causal identification; optimization settings, answer extraction, and training variance are competing explanations. The appropriate next experiments are cold-start reasoning SFT from a stronger teacher, easier verifiable tasks, reward curricula, multiple seeds, and repeating the pipeline at 500M and 1B parameters.

\section{What This Project Solves}

The project addresses four practical gaps that often block small-model research.

\paragraph{Budget transparency.}
Many small-model comparisons cite scores but hide the true gap in token and hardware budgets. Here the result is explicitly contextualized: \model\ is weaker than SmolLM-class systems but uses a tiny fraction of their reported training tokens. This makes the artifact useful for researchers who care about marginal gains per accessible GPU-hour.

\paragraph{Data hygiene under mixture construction.}
Modern small-model recipes increasingly mix web text, educational data, code, mathematics, synthetic textbooks, and SFT data. Without cross-source deduplication, repeated fragments and benchmark overlaps can silently dominate a small token budget. The Stage 4 pipeline treats deduplication, segment cleanup, and contamination checks as release gates rather than optional analysis.

\paragraph{Anti-forgetting post-training.}
Instruction tuning is not free at 135M. The SFT interpolation ablation shows that the best published checkpoint is not necessarily the final SFT checkpoint. Small models need regression-gated post-training, especially when one checkpoint is expected to serve both base-LM and simple instruction-following use cases.

\paragraph{Honest negative reasoning results.}
The RLVR experiment is useful precisely because it fails. It shows that verifiable rewards do not automatically create mathematical reasoning in a 135M model with weak initial GSM8K pass rate. This gives a concrete stop/go criterion before scaling the method to 500M or 1B.

\section{Limitations}

This report has several limitations.

\begin{itemize}
  \item \model\ is not state of the art among 135M models. SmolLM-135M and SmolLM2-135M are stronger on every measured task in our six-task comparison.
  \item The comparison is not architecture-controlled. Public baselines use different tokenizers, corpora, context lengths, and training schedules.
  \item The data cleaning ablation is incomplete. We have logged cleaning gates but not a matched no-dedup or relaxed-filter control run.
  \item The SFT improvements are small on base-LM benchmarks. Better chat evaluation and human preference evaluation are needed before claiming assistant quality.
  \item The contamination gate is threshold-based, not a formal proof. It reduces known overlap risk but cannot guarantee zero benchmark leakage.
  \item There is no repeat-seed training run. Reported benchmark differences should be interpreted cautiously.
  \item GPU-only energy estimates exclude host power, storage, networking, and idle time.
  \item Dataset-license and revision metadata must be verified against the release manifest before arXiv submission; this manuscript does not substitute for that audit.
\end{itemize}

\section{Threats to Validity}

\paragraph{Internal validity.}
Training stages and RLVR variants are single runs. The absence of replicated seeds prevents separating treatment effects from optimizer and sampling noise. SFT interpolation was selected on the same six-task aggregate that is reported, creating optimistic selection bias.

\paragraph{Construct validity.}
The unweighted mean combines heterogeneous tasks and metrics. It is a compact regression indicator, not a complete measure of language understanding, chat quality, safety, calibration, or factuality. GSM8K exact match is sensitive to answer formatting and extraction.

\paragraph{External validity.}
Findings are specific to one 135M architecture, one tokenizer, one L20 software stack, and the tested mixtures. Throughput measurements do not transfer directly to other accelerators or context-length distributions. The RLVR failure does not imply that RLVR fails for every 135M model or reward design.

\section{Future Work}

The highest-value next steps are controlled ablations rather than more ad hoc training. First, rerun short controlled pretraining variants that isolate cross-source deduplication, Stage 4 source mixture, 2K-to-8K curriculum, and SFT interpolation. Second, train to a 27B-token point to test whether this single-L20 recipe follows the expected small-model scaling curve. Third, repeat the RLVR pipeline at 500M and 1B only after adding cold-start distilled reasoning data and easier verifiable curricula. Fourth, publish bootstrap confidence intervals and per-task sample-level error buckets so that small benchmark gains are not overclaimed.

\section{Conclusion}

\model\ is best understood as a reproducible small-model systems result. It demonstrates that a single L20 GPU can support a complete 135M pretraining, data-curation, continued-pretraining, SFT, evaluation, and release loop. The final model is not a SmolLM replacement, but it reaches a meaningful fraction of modern 135M benchmark performance with a much smaller token and hardware budget, while documenting the exact data gates and post-training tradeoffs that make the result auditable. The most important scientific lesson is that data quality and regression control matter disproportionately at small scale, while reasoning RL appears to require either stronger cold-start competence or larger model capacity.

\section*{Artifact and Reproducibility Statement}

\begin{tabularx}{\linewidth}{lY}
\toprule
Artifact & Status \\
\midrule
Training config & Available in project configs for base, Stage 4, SFT, and RLVR runs \\
Training logs & 1,903 train-loss points and 38 validation-loss points for the 10B base run \\
Evaluation harness & EleutherAI lm-evaluation-harness 0.4.12, zero-shot full datasets \\
Raw eval outputs & Stored under \texttt{eval\_results/} for candidate and public baselines \\
Data gate summary & Stage 4 JSON records deduplication, quality, and contamination counters \\
Model release & Hugging Face repository \texttt{AliceYin/l20-edu-135m} \\
Randomness & Single training run per stage; no seed-based uncertainty estimate \\
Known gaps & Matched data ablations, repeat seeds, and validated chat/safety evaluation \\
\bottomrule
\end{tabularx}

The checkpoint is available at \url{https://huggingface.co/AliceYin/l20-edu-135m}. Reproduction should use the immutable configuration, dataset manifest, dependency lock, and raw evaluation outputs bundled with the project release. Before archival submission, every referenced artifact should be verified as publicly reachable without credentials and every data source should have an explicit revision and license record.

\section*{Broader Impact}

Lowering the compute barrier can broaden participation in language-model research and enable auditable local experiments. The same accessibility can also facilitate generation of misleading, biased, private, copyrighted, or unsafe text. This 135M checkpoint has not undergone deployment-grade safety evaluation, should not be used for consequential decisions, and can produce fluent but incorrect output. The release should include intended-use boundaries, upstream data notices, and a mechanism for reporting problematic memorization or outputs.

\bibliographystyle{plainnat}
\bibliography{references}

\end{document}